# Speech Synthesis with Neural Networks

**Orhan Karaali, Gerald Corrigan, and Ira Gerson**

*Motorola, Inc., 1301 E. Algonquin Road, Schaumburg, IL 60196*

*karaali@mot.com, corrigan@mot.com, gerson@mot.com*

### ABSTRACT

Text-to-speech conversion has traditionally been performed either by concatenating short samples of speech or by using rule-based systems to convert a phonetic representation of speech into an acoustic representation, which is then converted into speech. This paper describes a system that uses a time-delay neural network (TDNN) to perform this phonetic-to-acoustic mapping, with another neural network to control the timing of the generated speech. The neural network system requires less memory than a concatenation system, and performed well in tests comparing it to commercial systems using other technologies.

## 1.0 Introduction

### 1.1 Description of Problem

Text-to-speech conversion involves converting a stream of text into a speech waveform. This conversion process frequently includes the conversion of a phonetic representation of the text into a number of speech parameters. The speech parameters are then converted into a speech wave form by a speech synthesizer.

As the complexity and power of computers increase, means of communicating with them other than by traditional keyboard and display approaches becomes attractive. Speech dialog systems which include speech synthesis capabilities are considered as the ideal candidate for a number of computer user interfaces.

Also, in recent years the number of portable and automotive electronic products has increased greatly. These include communication devices, hand held computers, and automotive systems. While speech synthesizers can add value to portable products, their usage must not result in significantly higher product costs.

Concatenation and synthesis-by-rule have been the two traditional techniques used for speech synthesis (Klatt, 1987). Concatenative systems store patterns produced by an analysis of speech, diphones or demisyllables, and concatenate these stored patterns, adjusting their duration and smoothing transitions to produce speech parameters corresponding to the phonetic representation. One problem with concatenative systems is the large number of patterns that must be stored, usually on the order of a 1000 or more. Synthesis-by-rule systems are also used to convert phonetic representations into speech parameters. The synthesis-by-rule systems store target speech parameters for every possible phonetic representation. The target speech parameters are modified based on the transitions between phonetic representations according to a set of rules. The problem with synthesis-by-rule systems is that the transitions between phonetic representations are not natural, because the transition rules tend to produce only a few styles of transition. In addition, a large set of rules must be stored. Another major problem with the traditional synthesis techniques is the robotic speech quality associated with them.

### 1.2 Discussion of the use of networks

Market acceptance of portable equipment with speech synthesis capabilities, will depend on speech quality, functionality, and product costs. Since neural networks are trained from actual speech samples, they have the potential to generate more natural sounding speech than other synthesis technologies. While a rule based system requires generation of language dependent rules, a neural network based system is directly trained on actual speech data and, therefore, it is language independent, provided that a phonetically transcribed database exists for a given language. Concatenative systems can require several megabytes of data, which may not be feasible in a portable product. The neural network can generate a much more compact representation by eliminating the redundancies present in the concatenative approach. Recently, the advantages of the neural network based speech synthesis have resulted in a number of different research efforts in the area (Burniston and Curtis, 1994; Cawley and Noakes, 1993; Cohen and Bishop, 1994; Tuerk et al, 1991; Tuerk and Robinson, 1993; Weijters and Thole, 1993)

Figure 1 shows a block diagram of the text-to-speech system. The system has four major components. The first component converts text to a linguistic representation — phonetic and syntactic information — that can be converted to speech. The next component determines a segment duration for each phone generated by the text-to-linguistics component. The third component converts the phonetic and timing information into an acoustic representation. Finally, the synthesis portion of a vocoder is used to generate the speech. The analysis portion of the vocoder is used to generate data during the training phase of the neural network. The duration generation and phonetics-to-acoustics mapping are performed by neural networks.

## 2.0 System Description

The system has two basic modes of operation: training and testing. During the training phase both the duration and the phonetic neural networks are individually trained to produce their respective target values from the speech database. During the test phase, the database and the analysis section of the parametric speech vocoder are not used.

## 2.1 Database

A speech database is needed to train the neural networks used in the system. While several large databases are easily available, most are not suitable for this purpose. The most important requirement for the database is that it consist of a large number of speech samples from a single speaker. The speech also needs to be labelled both phonetically and prosodically for best results in training the neural networks. The collection and labelling of the database is described below.

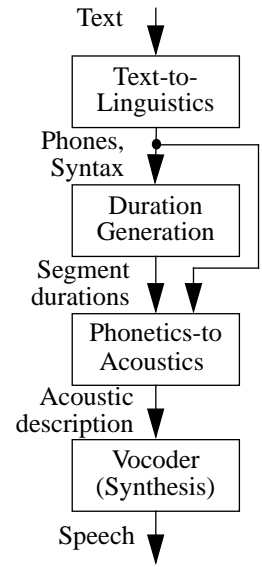

**Figure 1:** Text-to-speech system

The speaker that was used for the database is a 38 year old male speaker who has lived in both Florida and Chicago. The speech recording was performed on a DAT recorder using a close-talking microphone. The text consisted of 480 phonetically balanced sentences from the Harvard sentence list (IEEE Subcommittee on Subjective Measurements, 1969). In addition, 160 sentences recorded by this speaker for a different experiment were also used. The recordings were transferred digitally to a computer, where they were divided into separate files for each sentence. The sentences were then normalized so that each sentence file had the same average signal power across the non-silent parts of the file.

Several types of information must be marked in the speech database, to be used as the input information to the neural networks. The most obvious of this information is the sequence of phonetic segments that occur. Other information, described below, is used to determine the rhythm and intonation of the speech.

The speech sentences were labelled with the same phonetic labels and conventions used for the TIMIT database (Seneff and Zue, 1988). This is a variation on the ARPABET (Shoup, 1980), with stops labelled with the closure and release as separate phones. This provides useful information to the neural network, identifying whether a stop consonant was released, and indicating the precise start time of the release. Precise representations of the timing of events are useful to a neural network designed to minimize error on a frame-by-frame basis. Such networks should handle uncertainty in values better than they do uncertainty in timing.

The sequence of phonetic segments is not the only factor that is important in constructing synthetic speech. The segment durations and fundamental frequency variations, in particular, are affected by such things as syllable stress and syntactic boundaries. The syntactic labelling is designed to provide this information to the neural network. The syntactic labelling marks the start and end time of syllables, words, phrases, clauses, and sentences; the lexical stress (primary, secondary, or none) applied to each syllable in a word; whether each word is a function word (article, pronoun, conjunction, or preposition) or content word; and a "level" assigned to each word, based on a rule-based system for generating fundamental frequency (Allen et al, 1987).

Although the syntactic and lexical stress information is useful in determining prosodic variations in speech, that information does not fully determine those variations. A speaker will vary where he or she place a strong disjuncture in a sentence based on a variety of other factors, such as which information in an utterance is new information, as opposed to things the listener has already heard. For this reason, it may be more useful to label the speech marking the actual locations of such items as pitch accents on syllables, or strong disjunctures between words. One standard that has been developed for performing this labelling in English is the ToBI (Tone and Break Index) system (Silverman et al, 1992; Beckman and Hirschberg, 1994). This labelling system is also being utilized in the text-to-speech system.

## 2.2 Generation of segment durations from phonetic descriptions

One of the two tasks for which neural networks were used was to determine, from the sequence of phones to be uttered and from the syntactic and prosodic information, the duration of the phonetic segment associated with each occurrence of a phone.

The input to the network used mostly binary values, with categorical entries represented using 1-out-of n codes and some small integers values represented with bar codes. Input data used to represent a phonetic segment include the phone represented by the segment, its articulatory features, descriptions of the prominence of the syllable and the word containing the segment, and the proximity of the segment to any syntactic boundaries. The network was trained to generate the logarithm of the duration.

The network architecture for the duration network is shown in Figure 2. The network has two input streams (2 and 3) fed in through I/O blocks 1 and 2. (Stream 2 contains data that is fed in through a shift register to provide a context description for a given phone, while stream 3 contains input information that is only used in the generation of the duration for one particular phone. When the neural network is being used to generate durations, the I/O block 6 writes the results to output data stream 1. During training, block 6 reads the targets and generates the error value. Blocks 3, 4, and 5 are single-layer neural network blocks, while blocks 7, 8, and the recurrent buffer control the recurrence mechanism.

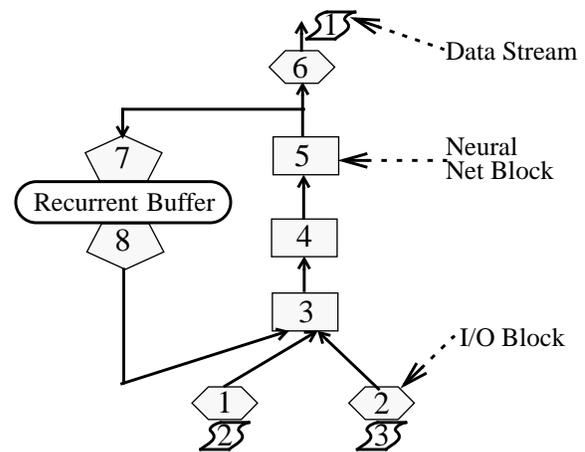

**Figure 2:** Duration network architecture

## 2.3 Generation of acoustic information from phonetic and timing information

The second neural network used in the system generates the speech parameters from the phonetic, syntactic, and timing data. More precisely, the network generates acoustic descriptions of ten millisecond frames of speech from a description of the phonetic context of the frame.

### 2.3.1 Network output - Coder

The neural network does not generate speech directly; that would be computationally expensive and unlikely to produce good results. The network generates frames of data for the synthesis portion of an analysis-synthesis style of vocoder. Most speech coding research has been focused on the problem of data compression; however, several of the neural network's requirements for a coder were not met by most speech compression techniques. In particular, it was important to encode speech as a vector of numbers for each frame such that each element of the vector had a well-defined value for every frame, and such that the error measure used by the neural network for training was appropriate. (i.e., if the neural network produced vectors for which the error measure relative to the training vectors was small, the quality of the speech produced by running these vectors through the synthesis portion of the coder would be good.) The weighted Euclidean distance which was used as the error criterion made it desirable that the coder not use binary output values, and that the meaning of vector elements not change based on the values of other vector elements.

The coder is a form of LPC vocoder, using line spectral frequencies to represent the filter coefficients, and a two-band excitation model. (Several different representations for the filter coefficients were tested, and the network performed well with line spectra.) The two-band excitation model is a variant of multi-band excitation, with a low-frequency voiced band, and a high-frequency unvoiced band. The boundary between these two bands is one of the coder's parameters. The fundamental frequency and power of the voice signal are the remaining parameters. The fundamental frequency is clamped to a high value during unvoiced frames.

### 2.3.2 Network input

The input to the phonetic network includes all the inputs used by the duration network with the addition of the timing information which was determined by the duration network. The network utilizes a number of different input coding techniques. The blocks 5, 6, 20, and 21 (see Figure 3) use 300 millisecond wide Time Delay Neural Network (TDNN) (Waibel et al, 1989) style input windows. The sampling of the windows are not uniform and the optimal sampling intervals are determined by analyzing the neural networks' usage of information from different parts of the TDNN windows. Blocks 6 and 20 process a set of features related to the input phonemes. Blocks 7 and 8 code duration and distance information both for phonemes and syntactic boundaries. The input data to the network is a mixture of binary values, 1-out-of n codes, and bar codes.

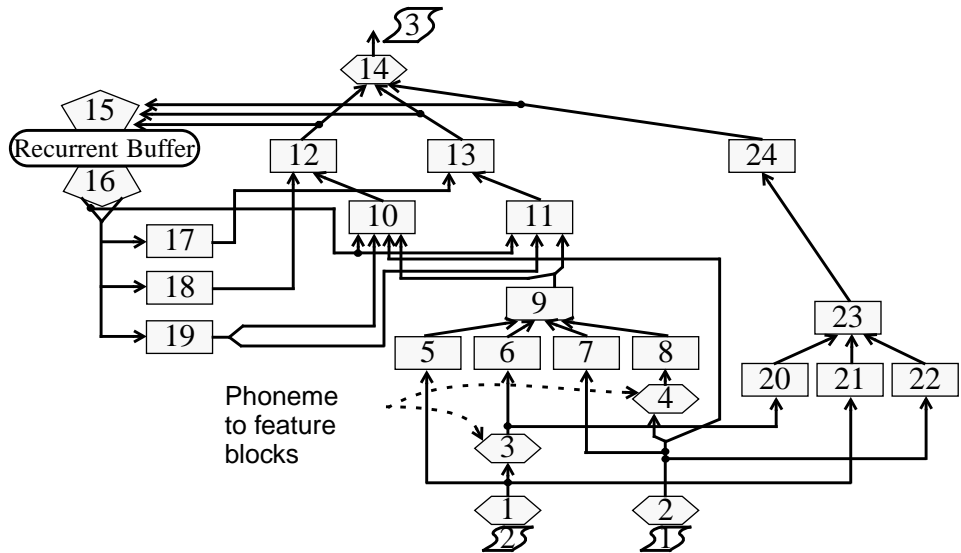

**Figure 3:** Phonetic network architecture

### 2.3.3 Network architecture

Determination of the proper network architecture requires training different networks and experimenting with them. Due to the size of the database and the complexity of the problem, the training time required with commercially available simulators turned out to be major bottleneck. Therefore, an in house neural network simulator was developed which reduced the training time from a couple of months to a few days and in the same time made it possible to implement different topologies with ease. Some of the neural network techniques and theories developed and proven for small networks and small data sets turned out to be not applicable for a problem of this complexity.

The final network topology combines the advantages of TDNN, recurrent, and modular networks along with empirically evolved techniques. Figure 3 is the block diagram of the present topology where the hexagonal blocks are I/O or user written subroutines and the rectangular blocks are the neural network modules. Neural networks blocks were trained using backpropagation. Modularization of the network was done by hand using speech domain expertise.

The network is trained with decreasing learning rate and momentum in a novel mixture of sequential and random training modes (Karaali, 1994). The trained network requires less than 100 Kilobytes of eight bit quantized weights which compares very favorably with the few megabytes required for concatenation systems.

### 3.0   Performance of the system

### 3.1   Speech quality and intelligibility

Figure 4 contains spectrograms of natural speech and the synthesized speech generated by the system. (The synthesized sentences shown were generated using a version of the system that did not use the ToBI labeling.) To facilitate comparison of the sounds, two different versions of the synthesized sentence were generated. In the first synthesized example, the acoustic information was generated from the phonetic and timing information using the actual segment durations from natural speech, illustrating the behavior of just the phonetic neural network. In the second example, the timing information for the phonetic neural network was generated by the duration neural network.

Independent testing results (Nusbaum et al, 1995) have shown that the neural network system performed much better than other commercial speech synthesizers in the area of speech acceptability which is an overall assessment. This is understandable since the neural network system generates very natural human-like speech. In the area of word level intelligibility, the neural network was not the best system. While there is a lot of room for improving word intelligibility, one of the reasons for the lower score in this test might be due to the lack of single word speech samples in the training database.

| System | Acceptability | Segmental Intelligibility |
|---|---|---|
| Human | Not tested | 86% |
| **Motorola** | **4.3** | **55%** |
| DECTalk | 3.5 | 71% |
| TrueTalk | 3.5 | 72% |
| PlainTalk | 2.3 | 60% |

**Table 1:** Comparison of Speech Synthesizers

### 3.2 Real time implementation

The system was originally developed on a Sun SPARCstation platform using the ANSI C language. Recently, it has been ported to the Power Macintosh 8500/120. The existence of the fast multiply and add instruction on the PowerPC chip makes it possible to run the synthesizer in real time.

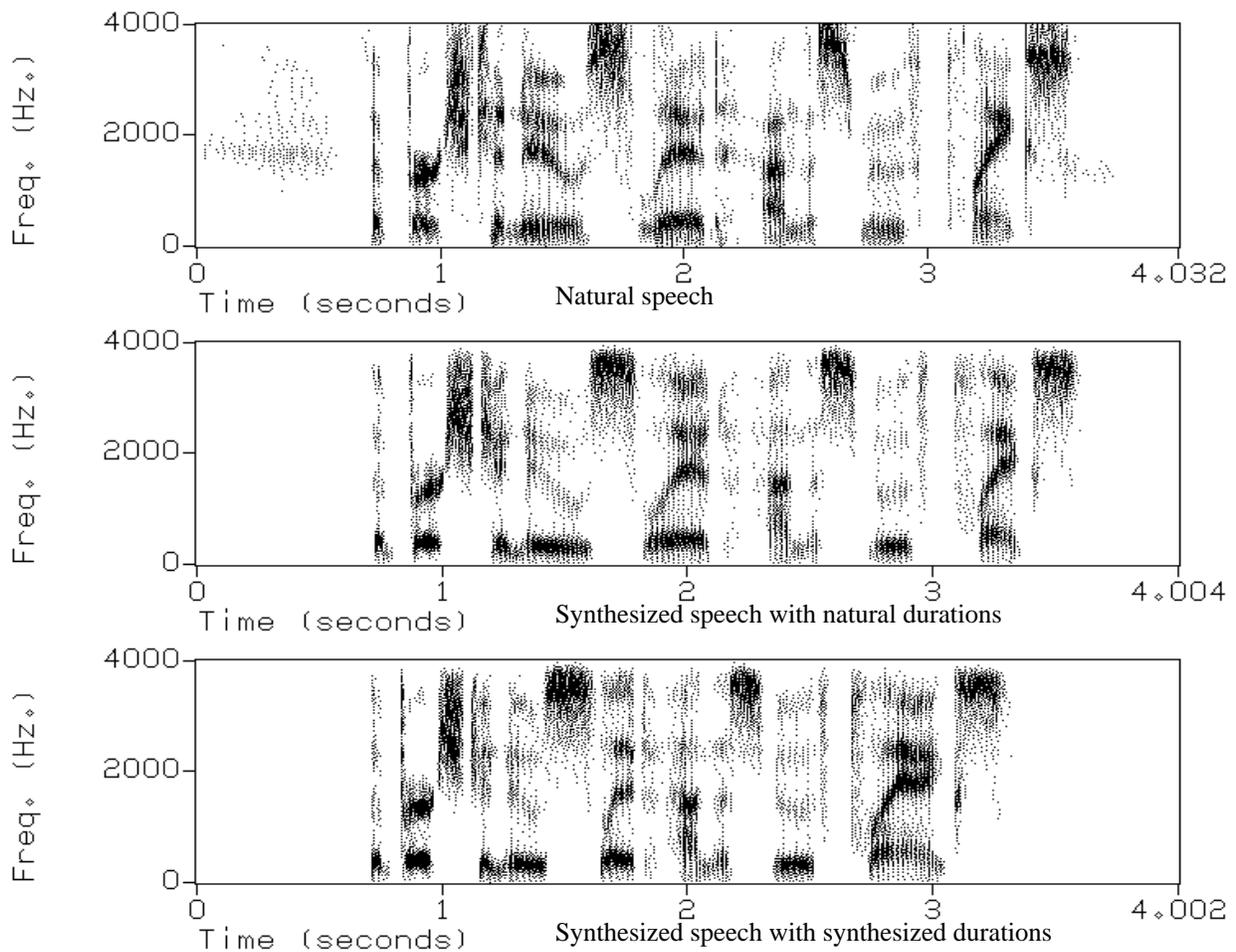

**Figure 4:** Comparison of natural speech, synthesized speech with natural durations, and synthesized speech with synthetic durations for the sentence: "The birch canoe slid on the smooth planks."

## 4.0 Conclusion

The neural network approach to speech synthesis offers the benefits of language portability, natural sounding speech, and low storage requirements. The results from the acceptability experiments indicate that neural network based text-to-speech systems have the potential to provide better voice quality than traditional approaches, but some improvement in the system is still desirable. The database needs to be expanded to include more intonational variation than is present in the current version, as well as to include as many phonetic contexts as is possible. In particular, the database should include examples of short utterances — single words and brief phrases — and long, paragraph-length, utterances. Improvements in the coder, network architecture, and training methods may also be possible.